\documentclass[journal,twoside]{IEEEtran}
\usepackage{amsmath,amsfonts}
\usepackage{algorithmic}
\usepackage{array}
\usepackage[caption=false,font=normalsize,labelfont=sf,textfont=sf]{subfig}
\usepackage{textcomp}
\usepackage{stfloats}
\usepackage{url}
\usepackage{verbatim}
\usepackage{graphicx}
\usepackage{amsmath,amssymb,amsfonts}
\usepackage[ruled,linesnumbered]{algorithm2e}
\usepackage{graphicx}
\usepackage{subfig}
\usepackage{textcomp}
\usepackage{epstopdf}
\usepackage{xcolor, color}
\usepackage{float}
\usepackage{multirow}
\usepackage{booktabs}
\usepackage{enumitem}
\usepackage{makecell}
\usepackage{bm}
\usepackage{amsfonts}
\usepackage{amssymb}
\makeatletter

\newcommand{\Rmnum}[1]{\expandafter\@slowromancap\romannumeral #1@}
\makeatother
\usepackage{upgreek}
\usepackage{setspace}
\usepackage{threeparttable}
\usepackage{soul}
\usepackage{multirow}
\usepackage{multicol}
\usepackage[hidelinks, colorlinks=true, citecolor=blue, linkcolor=blue, urlcolor=blue]{hyperref}
\usepackage[capitalize]{cleveref}
\usepackage{nomencl}
\usepackage{ifthen}
\usepackage{soul}
\usepackage{tcolorbox}
\makenomenclature
\soulregister\cite7 
\soulregister\citep7 
\soulregister\citet7 
\soulregister\ref7 
\soulregister\pageref7 
\soulregister{\underline}7
\soulregister{\Rmnum}7
\soulregister\cref7

\def\BibTeX{{\rm B\kern-.05em{\sc i\kern-.025em b}\kern-.08em
    T\kern-.1667em\lower.7ex\hbox{E}\kern-.125emX}}
\usepackage{balance}
\makeatletter
\renewcommand\subsubsection{\@startsection{subsubsection}{3}{0\parindent}%
  {0.1ex plus 0.1ex minus 0.1ex}%
  {0.1ex}%
  {\normalfont\normalsize\itshape}%
}

\usepackage{amsmath}
\begin{document}
\title{Scale-Adaptive Power Flow Analysis with Local Topology Slicing and Multi-Task Graph Learning}
\author{Yongzhe Li, Lin Guan, \textit{Member, IEEE}, Zihan Cai, Zuxian Lin, Jiyu Huang, Liukai Chen
\thanks{This work was supported in part by the Smart Grid-National Science and Technology Major Project of China (2025ZD0804900) and the National Natural Science Foundation of China (No. U22B6007).\textit{(Corresponding author: Lin Guan.)}

Y. Li, L. Guan, Z. Cai and Z. Lin are with the School of Electric Power, South China University of Technology, Guangzhou 510641, China (e-mail: 10706719873@qq.com; lguan@scut.edu.cn; epc\_zihan@mail.scut.edu.cn; 2660910069@qq.com)

J. Huang is with the CSG Energy Development Research Institute Co., Ltd., Guangzhou, 510663, China.

L. Chen is with the Electric Power Research Institute of China Southern Power Grid Company Limited, Guangzhou 510663, China.
}}

\maketitle

\begin{abstract}
Developing deep learning models with strong adaptability to topological variations is of great practical significance for power flow analysis. To enhance model performance under variable system scales and improve robustness in branch power prediction, this paper proposes a \underline{S}cale-\underline{a}daptive \underline{M}ulti-task \underline{P}ower \underline{F}low \underline{A}nalysis (SaMPFA) framework. SaMPFA introduces a Local Topology Slicing (LTS) sampling technique that extracts subgraphs of different scales from the complete power network to strengthen the model's cross-scale learning capability. Furthermore, a Reference-free Multi-task Graph Learning (RMGL) model is designed for robust power flow prediction. Unlike existing approaches, RMGL predicts bus voltages and branch powers instead of phase angles. This design not only avoids the risk of error amplification in branch power calculation but also guides the model to learn the physical relationships of phase angle differences. In addition, the loss function incorporates extra terms that encourage the model to capture the physical patterns of angle differences and power transmission, further improving consistency between predictions and physical laws. Simulations on the IEEE 39-bus system and a real provincial grid in China demonstrate that the proposed model achieves superior adaptability and generalization under variable system scales, with accuracy improvements of 4.47\% and 36.82\%, respectively.
\end{abstract}

\begin{IEEEkeywords}
Power flow analysis, graph learning, cross-scale adaptation, local topology slicing, multi-task learning.
\end{IEEEkeywords}

\section{Introduction}
\label{introduction}
\IEEEPARstart{D}{EEP} learning-based power flow analysis (DL-based PFA) has emerged as an effective approach to rapidly and accurately evaluate the massive operating scenarios induced by the uncertainty of renewable energy sources \cite{ref1}. DL-based PFA models typically take bus powers and generator voltages as inputs and generate bus voltage magnitudes and phase angles through nonlinear neural mappings. This direct mapping eliminates the iterative process and complex matrix operations required in traditional numerical methods, offering fast computation \cite{ref2,ref3} and freedom from convergence issues \cite{ref4}. As DL-based PFA research progresses, the practical applicability of such models in real-world power systems has become a major focus.

Real-world power systems operate in complex and variable scenarios, accompanied by topological changes such as line switching during maintenance and the integration of new substations and transmission lines during system expansion. Successful application of the DL-based PFA technique in real-world power grids requires further improvements in two aspects: handling varying operation scenarios with high accuracy \cite{ref5}, and addressing changes in grid scale and structure with strong adaptation and generalization.

\textbf{1) Adaptability to system-scale variations}

Foundational work of enhancing model adaptability to topology change includes incorporating power network parameters such as bus self-admittance \cite{ref6,ref7} and branch indices \cite{ref8} into the input, and applying graph neural networks (GNNs) \cite{ref9,ref10} to interpret power system topologies. Subsequent improvements, such as enlarging the model receptive field \cite{ref11, ref12, ref13, ref14} and designing physics-based graph convolution kernels \cite{ref15}, further increase the robustness of model performance under varying topologies. Inspired by large language models (LLMs), some studies combined GNNs with Transformer architectures and employed pre-training and fine-tuning to achieve transferable performance across multiple PFA tasks \cite{ref16}. Nevertheless, existing studies are still limited to cases where line switching is allowed while the number of buses remains fixed.

The ultimate goal of DL-based PFA is to build a general model applicable to power system expansion or even to multiple systems with different topologies and bus scales. Existing studies have shown that although the operation of GNNs is theoretically independent of network size, their accuracy often drops sharply or even fails when the number of buses changes \cite{ref17}. From the perspective of artificial intelligence, the fundamental reason lies in the distributional shift of graph statistics \cite{ref18}. There are research works in other fields towards this issue, such as generating synthetic graphs of varying scales to enhance cross-scale adaptability in social network applications \cite{ref19}, adopting subgraph sampling to improve performance on large graphs \cite{ref20}, etc.. 

Power systems possess inherent local separability that any local subgraph can be regarded as an independent sample for learning. Therefore, extracting subgraphs of different scales from the original power system may be an effective way to enlarge the training set size and diversity, enable the model to learn cross-scale power flow laws, and improve the scale adaptability. However, in power flow analysis, subgraph sampling inevitably leads to inconsistencies in the selection of slack buses, resulting in unaligned reference frames for bus phase angles and increasing the difficulty of learning phase angle mapping.

\textbf{(2) Accuracy in real-world power systems with multiple voltage levels}

Existing DL-based PFA models are typically designed to predict bus voltage magnitudes and phase angles, and then the branch power values are computed using power flow equations \cite{ref16}. Under this framework, there have been fruitful research works on the enhancement of model accuracy. A widely adopted skill is to embed the physical principles into neural network training \cite{ref1,ref5, ref21, ref22, ref23, ref24, ref25}. Efforts have been made to incorporate Kirchhoff's current law (KCL) \cite{ref22}, AC power flow equations \cite{ref23}, and operational constraints \cite{ref24} into loss functions, encouraging the model to follow physical laws while improving accuracy. In \cite{ref25}, a second-order cone relaxation is employed to linearize AC equations, simplifying the learning process and improving precision. 

Nonetheless, since most of the studies are tested on standard benchmark systems with idealized impedance parameters, the error amplification phenomenon that exists in real-world power systems with very small impedance branches hasn't been revealed. Such ill-impedance branches result from three-winding transformers, bus switches, etc., and are widely present in real-world power system models, which include multiple voltage levels and the substation bus layouts. During power flow calculation of these low impedance branches, small voltage prediction errors in the DL-based PFA model can be significantly amplified and result in large branch power deviations. 

To address the above challenges, this paper proposes a Scale-adaptive Multi-task Power Flow Analysis framework (SaMPFA) to enhance the practical applicability of DL-based PFA. SaMPFA can accommodate power system scale expansion through a Local Topology Slicing (LTS)-based training augmentation and the phase-reference-free design. Also, the proposed model directly predicts all the branch powers, thereby eliminating the error amplification risk in the branch power computation. The main contributions of this paper are summarized as follows: 

1) A Reference-free Multi-task Graph Learning (RMGL) model is developed for the DL-based PFA task. Instead of direct phase angle prediction, the RMGL scheme employs the multi-task design to jointly predict the bus voltage and the branch power, which effectively avoids error amplification in the phase-angle-based branch power calculation. Then, the bus phase angle difference can be calculated according to branch-related physical equations. This design also achieves adaptability to the slack bus variations and therefore can support the samples generated by LTS.

2) The LTS-based data augmentation method is proposed, which collects subgraphs of various scales from the power network and populates the training dataset. LTS significantly enriches the topology pattern and scale diversity of samples, therefore improving the adaptability and generalization of the trained PFA model.

3) A group of physics-guided constraints is incorporated to guide the model, including the physical laws of branch angle differences and power losses, which ensures the reliability of the results and improves the accuracy.

4) Comprehensive experiments on a real-world power system with scenarios up to 10\% network size variation are carried out. Domain test demonstrates that the proposed model exhibits excellent cross-scale adaptability and generalization capability.

The remainder of this paper is organized as follows. \cref{section 2} introduces the challenges and framework of SaMPFA. \cref{section 3} and \cref{section 4} describe the structure and training strategy of the RMGL model. \cref{section 5} presents the case studies, and \cref{section 6} concludes the paper. 

\section{Framework Design}
\label{section 2}

\subsection{Key Challenges in Cross-Scale Adaptability}
\label{section 2.1}
\subsubsection{Distribution Shifts Caused by Scale Variations}
\label{section 2.1.1}
\begin{figure}[t]
\centerline{\includegraphics[width=0.93\columnwidth]{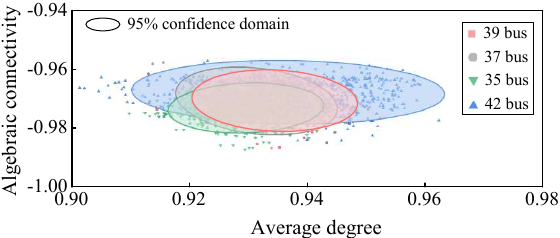}}
\vspace{-0.35cm}
\caption{Distributions of graph characteristics under different bus counts.}
\label{fig1}
\vspace{-0.2cm}
\end{figure}

To quantify the impact of variable system scale on graph statistics, the average node degree and algebraic connectivity \cite{ref26} are selected as the key indicators for the graph topology. The adjacency matrix $\boldsymbol{A}$ is constructed based on the branch reactance, and the Laplacian matrix is obtained as $\boldsymbol{L}=\boldsymbol{D}^{-1/2}\boldsymbol{A}\boldsymbol{D}^{-1/2}$, where $\boldsymbol{D}$ denotes the degree matrix of $\boldsymbol{A}$. The average degree is the mean of the degree matrix of $\boldsymbol{L}$, and the algebraic connectivity is the second smallest eigenvalue of $\boldsymbol{L}$. In \cref{fig1}, random topology changes are carried out on the IEEE 39-bus system by new station expansion, existing generator shutdown, and line outages. A total of 4,000 network samples with different bus amounts are thus generated, and their distribution in the average node degree and algebraic connectivity feature space is shown in \cref{fig1}.

It can be observed that the statistical properties of the network exhibit evident shifts related to system bus size. This may explain, from one perspective, why a well-trained PFA model experiences a deterioration in performance when the power grid scale is expanded.

The most straightforward way to enhance the scale adaptability of DL-based PFA models is to augment the training dataset, increase its diversity in both network size and topology. However, a great change in power grid size will lead to a new challenge caused by the slack bus selection. In power flow calculation, a change of the slack bus not only affects the power distribution, but also shifts all the bus phase angles uniformly \cite{ref27}. From the perspective of the DL model, if the model outputs include bus phase angles, it should learn how slack bus selection affects all the phase angle values, which significantly increases the difficulty in integrated training of variant power system scales. 

\subsubsection{Error Amplification in Branch Power Computation}
\label{section 2.1.2}
Existing DL-based PFA models usually follow classic power flow calculation definitions and select bus voltage magnitudes and angles as the direct output. The branch powers are then computed by the following power flow equation \cite{ref2}:
\begin{equation}
  \begin{split}
    {\overset{\rightharpoonup}{S}_{L,ij}} = &{\overset{\rightharpoonup}V _i}{(({\overset{\rightharpoonup}V _i} - {\overset{\rightharpoonup}V _j}){\overset{\rightharpoonup}y _{L,ij}})^*} \\
    = &\overset{\rightharpoonup}y _{L,ij}^*(V_i^2 - {V_i}{V_j}{e^{j{\theta _{ij}}}})
  \end{split}
  \label{eq1}
\end{equation}
where ${\overset{\rightharpoonup}S _L}$ is the complex branch power. ${\overset{\rightharpoonup}V _i} = {V_i}\angle {\theta _i}$ and ${\overset{\rightharpoonup}V _j} = {V_j}\angle {\theta _j}$ are the voltages at bus $i$ and $j$, and ${\theta _{ij}} = {\theta _i} - {\theta _j}$ is their angle difference. ${\overset{\rightharpoonup}y _L} = {g_L} + j{b_L}$ is the branch admittance. The superscript * indicates the complex conjugate.

Since DL models cannot perfectly predict all bus voltages, errors in voltages and angles can propagate through (\ref{eq1}). To evaluate this, we linearize (\ref{eq1}) at an operating point $(V_i, V_j, \theta_{ij})$:
\begin{equation}
  \begin{split}
\partial {S_{L,ij}} = & \overset{\rightharpoonup}y _{L,ij}^*{e^{j{\theta _{ij}}}}(2{V_i}\partial {V_i}\\
             &- {e^{j{\theta _{ij}}}}({V_j}\partial {V_j} + {V_i}\partial {V_j} + j{V_i}{V_i}\partial {\theta _{ij}}))\\
         = &{K_1}\overset{\rightharpoonup}y _{L,ij}^*\partial {V_i} - {K_2}\overset{\rightharpoonup}y _{L,ij}^*\partial {V_j} - j{K_3}\overset{\rightharpoonup}y _{L,ij}^*\partial {\theta _{ij}}
  \end{split}
  \label{eq2}
\end{equation}
\begin{equation}
  {K_1} = 2{V_i} - {V_j}{e^{j{\theta _{ij}}}},{K_2} = {V_i}{e^{j{\theta _{ij}}}},{K_3} = {V_i}{V_i}{e^{j{\theta _{ij}}}}
  \label{eq3}
\end{equation}

\begin{figure}[t]
\centerline{\includegraphics[width=0.92\columnwidth]{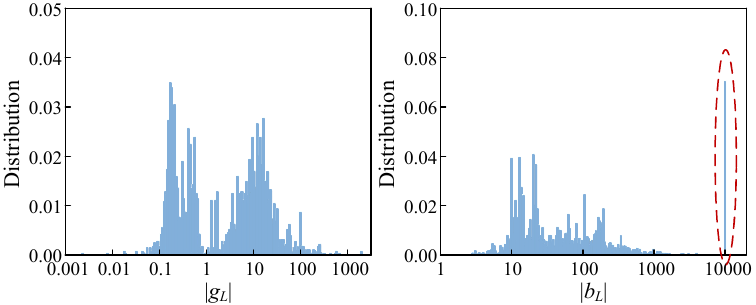}}
\vspace{-0.35cm}
\caption{Per-unit admittance distribution of branches in a real-world power system.}
\vspace{-0.2cm}
\label{fig:admittance}
\end{figure}

Since ${V_i} \approx {V_j} \approx 1$, the error coefficients are proportional to $\left| {{K_1}} \right| \approx \left| {{K_2}} \right| \approx \left| {{K_3}} \right| \approx 1$. This indicates that the branch power error is approximately proportional to the errors in bus voltages and angle differences, with a ratio of about $\left| {{{\overset{\rightharpoonup}y }_{L,ij}}} \right|$. As shown in \cref{fig:admittance}, real-world power system models usually contain branches with very low impedance, such as the inclusion of the bus tie switches or the medium-voltage side of three-winding transformers. Since the admittances of such branches can reach $10^3$ to $10^4$ p.u., even small voltage errors may be amplified, leading to branch power errors greater than 1 p.u..

Towards the above two challenges, a novel design of the DL-based PFA model is proposed in this paper. Direct outputs of the DL-based model are set to be the branch powers, instead of the bus phase angle, which avoids the aforementioned error amplification. Then the phase angle difference of any two buses is calculated by various forms of (\ref{eq1}) according to the branch power. Given a phase reference point (the slack bus), the bus phase angles can be easily obtained, which effectively eliminates the difficulty in phase angle prediction caused by the random selection of the slack bus in power grids of different scales. 

\subsection{Framework of SaMPFA}
\label{section 2.2}
\subsubsection{Architecture}
\label{section 2.2.1}
\begin{figure}[t]
\centerline{\includegraphics[width=1.0\columnwidth]{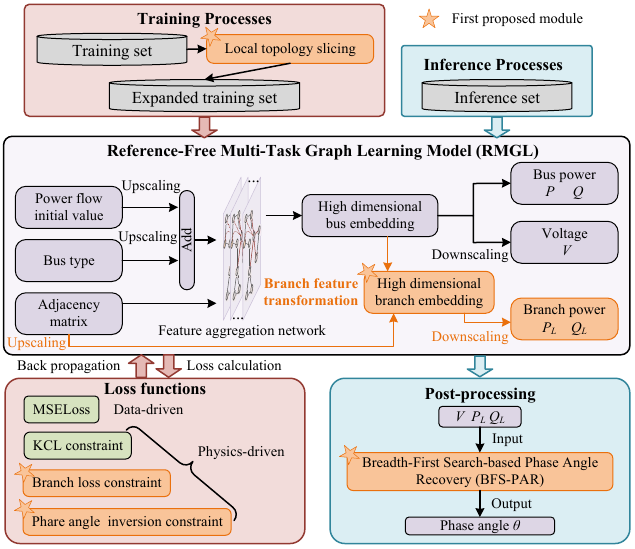}}
\vspace{-0.35cm}
\caption{Framework of SaMPFA.}
\label{fig:framework}
\vspace{-0.1cm}
\end{figure}

\cref{fig:framework} illustrates the framework of the so-called Scale-adaptive Multi-task Power Flow Analysis (SaMPFA) model proposed in this paper.

The kernel of the scheme is the Reference-free Multi-task Graph Learning model (RMGL) in the middle block. In a power system with $N$ buses and $E$ branches, the input of RMGL ${{\boldsymbol{X}}_{in}} \in {\mathbb{R}^{N \times 7}}$ includes the initial values of active power, reactive power, and voltage magnitudes at all buses, as well as the reactive power bounds and self-admittance. The weighted adjacency matrix $\boldsymbol A$ encodes the branch admittances and tap ratios. To avoid the learning difficulty induced by slack bus changes, phase angles are not explicitly included in either inputs or outputs. The RMGL performs two tasks. The first is a bus-level prediction, which outputs the active power, reactive power, and voltage magnitude of each bus, with the output denoted as ${\widetilde {\boldsymbol{X}}_{out}} \in {\mathbb{R}^{N \times 3}}$. The second is a branch-level prediction that outputs the active and reactive power of all branches, with the output denoted as ${\widetilde {\boldsymbol{H}}_{out}} \in {\mathbb{R}^{2E \times 2}}$. After feature extraction, the high-dimensional representations of the two terminal buses of each branch are fused and mapped to the branch power through a downscaling network, which significantly enhances numerical robustness.

During training, we adopt Local Topology Slicing (LTS) to sample subgraphs with varying sizes and connection patterns from the full grid. This strategy enlarges the dataset and enriches topological diversity, enabling the model to learn shared physical regularities across different system scales.

The loss function combines data fidelity and physics guidance. In addition to the mean squared loss (MSELoss) and KCL constraint, we introduce an angle restoration constraint to guide learning of branch angle differences, and a branch loss constraint to enforce power transfer and loss relationships. The joint optimization maintains prediction accuracy while preserving physical consistency.

\subsubsection{Phase Angle Inference and Application}
\label{section 2.2.2}
At inference time, the model takes an operating scenario as input and outputs bus active power, reactive power, voltage magnitudes, and branch power for all branches. The phase angle difference across each branch is computed using (\ref{eq4}).
\begin{equation}
  {\theta _{ij}} = \arctan \frac{{{b_{L,ij}}{P_{L,ij}} + {g_{L,ij}}{Q_{L,ij}}}}{{{g_{L,ij}}{P_{L,ij}} - {b_{L,ij}}{Q_{L,ij}} - V_i^2(g_{L,ij}^2 + b_{L,ij}^2)}}
  \label{eq4}
\end{equation}
where $P_{L,ij}$ and $Q_{L,ij}$ are the active and reactive power of branch $i-j$.

Based on (\ref{eq4}), we propose Breadth First Search-based Phase Angle Recovery (BFS-PAR) to reconstruct bus voltage angles over the entire network. The power system is represented as an undirected graph ${\cal G}({\cal B},{\cal E})$, where ${\cal B}$ and ${\cal E}$ denotes the set of buses and branches. Given a slack bus $b_{slack}$ and its assigned angle $\theta_{slack}$, a BFS strategy is applied to iteratively recover the voltage angles of all buses, as illustrated in \cref{code1}.

\begin{algorithm}[t]
\caption{BFS-based Phase Angle Recovery}
\label{code1}
\KwIn{Bus voltage magnitudes $\boldsymbol V$, branch active powers $\boldsymbol{P}_L$, branch reactive powers $\boldsymbol{Q}_L$, slack bus $b_{slack}$ and its reference angle $\theta_{slack}$}
\KwOut{Phase angles $\boldsymbol{\theta}$}
Compute angle differences $\theta_{ij}$ across branches by (\ref{eq4})\\
Initialize a queue ${\cal Q} \leftarrow \{ {b_{slack}}\} $, and the recovered bus set ${{\cal B}_{visited}} \leftarrow \{ {b_{slack}}\} $\\
\textbf{while} $\left| {{{\cal B}_{visited}}} \right| \ne \left| {\cal B} \right|$ \textbf{do} \\
\quad Dequeue a bus $b_q$ from $\cal Q$\\
\quad Search the neighbor set ${\cal N}({b_q})$\\
\quad \textbf{for} ${b_n} \in {\cal N}({b_q})\backslash {{\cal B}_{visited}}$ \textbf{do} \\
\quad \quad Recover the angle at $b_n$: ${\theta _n} = {\theta _q} - {\theta _{nq}}$\\
\quad \quad Add $b_n$ to ${{\cal B}_{visited}}$ and enqueue to $\cal Q$\\
\quad \textbf{end for}\\
\textbf{end while}
\end{algorithm}

\section{Model Design Methodology}
\label{section 3}
This section presents the design of the proposed RMGL model. \cref{section 3.1} describes the structure and computational process of RMGL. \cref{section 3.2} details the design of the feature extraction network. \cref{section 3.3} introduces the multi-task design scheme that mitigates the risk of error amplification.

\subsection{Structure of RMGL}
\label{section 3.1}

\begin{figure}[t]
\centerline{\includegraphics[width=1.0\columnwidth]{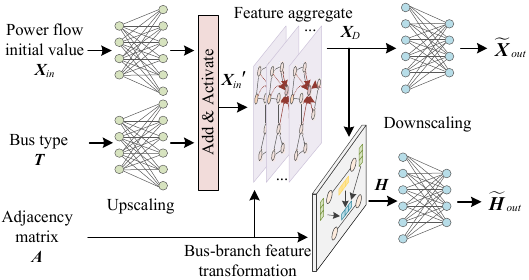}}
\vspace{-0.35cm}
\caption{Structure of RMGL.}
\label{fig:RMGL}
\vspace{-0.2cm}
\end{figure}

To achieve adaptability and generalization under varying system scales, RMGL must be able to recognize and handle changes in the number of buses and branches. Given the strong performance of Graph Transformers \cite{ref28} in topology recognition and structural generalization, RMGL is built upon this architecture. Its structure is shown in \cref{fig:RMGL}.

Since the self-attention mechanism in Transformers requires inputs of fixed length, RMGL predefines a maximum number of buses, denoted as $N_{\max}$. This design is similar to the maximum token length used in LLMs. When the actual system size $N$ satisfies $N\leq N_{\max}$, the model can operate normally. Therefore, $N_{\max}$ must be set large enough to cover potential future system scales. For cases where $N < N_{\max}$, the remaining $N_{\max}-N$ virtual buses are filled with arbitrary values as placeholders in the input features.

It should be noted that changes in the number of buses modify the arrangement of the input feature matrix $\boldsymbol{X}_{in}$. To reflect the type of each bus and the status of its associated variables, RMGL introduces an additional bus type matrix $\boldsymbol{T}$, in addition to the input feature matrix $\boldsymbol{X}_{in}$ and adjacency matrix $\boldsymbol A$ The encoding scheme is defined as follows:
\begin{enumerate}[label=$\bullet$]
  \item PQ bus: $T_i = [1,1,0]$, indicating that active power $P$ and reactive power $Q$ are known, and voltage magnitude $V$ is the target variable.
  \item PV bus: $T_i = [1,0,2]$, where $P$ and $V$ are known, and $Q$ is to be solved. The value 2 for $V$ indicates that its status may change if it switches to the PQ bus due to reactive power limit violations.
  \item Slack bus: $T_i = [0,0,1]$, indicating that voltage magnitude $V$ is known, and both $P$ and $Q$ are to be predicted.
  \item Virtual bus: $T_i = [0,0,0]$, indicating that the bus is a virtual placeholder whose features are masked in subsequent computations.
\end{enumerate}

During the feedforward process, the initial value $\boldsymbol{X}_{in}$ and the bus type matrix $\boldsymbol T$ are first projected into a high-dimensional feature space, yielding the input embeddings $\boldsymbol{X}_{in}^\prime$. 
\begin{equation}
  {\boldsymbol{X}_{in}^\prime}  = \sigma ({{\boldsymbol{X}}_{in}}{{\boldsymbol{W}}_{in}} + {\boldsymbol{T}}{{\boldsymbol{W}}_T})
  \label{eq5}
\end{equation}
where ${{\boldsymbol{W}}_{in}} \in {\mathbb{R}^{7 \times {d_D}}}$ and ${{\boldsymbol{W}}_T} \in {\mathbb{R}^{3 \times {d_D}}}$ are learnable projection matrices for the initial features and type matrix, $\sigma ( \cdot )$ is a nonlinear activation function, and $d_D$ denotes the dimension of the high-dimensional embedding.

Subsequently, an $M$-layer Masked Graph Transformer (MGT) is used to aggregate features among buses, yielding the high-dimensional power flow representation $\boldsymbol{X}_D$:
\begin{equation}
  {{\boldsymbol{X}}_D} = f_{{\rm{MGT}}}^M({{\boldsymbol{X}}_{in}^\prime} ,{\boldsymbol{A}})
\end{equation}

Finally, the bus embedding $\boldsymbol{X}_D$ is passed through a fully connected neural network (FCNN) to generate bus-level power flow predictions. Meanwhile, branch power flows are obtained by extracting branch-level representations $\boldsymbol{H}$ via a bus-branch feature transformation module, followed by a separate FCNN for output projection.

\subsection{Feature Propagation via Masked Graph Transformer}
\label{section 3.2}
The Graph Transformer architecture integrates the local topological modeling capability of GNNs with the global feature extraction capability of Transformers. Specifically, the Transformer module employs a multi-head attention (MHA) mechanism to capture dependencies between arbitrary pairs of buses, enhancing generalization to large-scale topologies.

The GNN module aggregates local neighborhood information to extract the structural characteristics of the power grid, enabling the model to handle various network configurations.

Denoted $\boldsymbol{X}_{GT}$ and $\boldsymbol{X}_{GT}^\prime$ as the bus embeddings before and after aggregation through the Graph Transformer. The aggregation process is expressed as:
\begin{equation}
  {{\boldsymbol{X'}}_{GT}} = {f_{MHA}}({{\boldsymbol{X}}_{GT}}) + {f_{GNN}}({{\boldsymbol{X}}_{GT}},{\boldsymbol{A}}) + {{\boldsymbol{X}}_{GT}}
  \label{eq7}
\end{equation}

Since GNNs are inherently capable of managing variations in the number of buses and branches, the Graph Attention Network (GAT) is adopted as the GNN module, whose feed-forward function is defined in \cite{ref29}.

Because the input includes virtual buses that should not participate in feature propagation, a masking mechanism is introduced within the MHA to block them. The attention computation is defined as follows:
\begin{equation}
  {{\boldsymbol{\alpha }}^{(k)}} = \frac{{({{\boldsymbol{X}}_{GT}}{\boldsymbol{W}}_Q^{(k)}){{({{\boldsymbol{X}}_{GT}}{\boldsymbol{W}}_K^{(k)})}^{\rm{T}}}}}{{\sqrt {{d_D}/K} }}
  \label{eq8}
\end{equation}
\begin{equation}
  {M_{ij}} = \left\{ {\begin{array}{*{20}{l}}
{ - \infty }&{i/j \in {\cal P}}\\
0&{{\rm{else}}}
\end{array}} \right.
\label{eq9}
\end{equation}
\begin{equation}
  {f_{MHA}}({{\boldsymbol{X}}_{GT}}) = \parallel _k^K{\rm{softmax}}({{\boldsymbol{\alpha }}^{(k)}} + {\boldsymbol{M}})({{\boldsymbol{X}}_{GT}}{\boldsymbol{W}}_V^{(k)})
  \label{eq10}
\end{equation}
where $K$ is the number of attention heads. $\boldsymbol{\alpha }^{(k)}$ denotes the attention weight matrix for the $k$-th head. The mask matrix $\boldsymbol M$ is configured such that if either bus $i$ or $j$ is a padded bus, the corresponding entry is set to $-\infty$, ensuring a zero-attention score. ${\boldsymbol{W}}_Q^{(k)}$, ${\boldsymbol{W}}_K^{(k)}$, and ${\boldsymbol{W}}_V^{(k)}$ represent the query, key, and value projection matrices for the $k$-th head.

Finally, the aggregated features are passed through a feed-forward network (FFN) \cite{ref30} with nonlinear activation to complete the feature update.

After stacking multiple MGT layers, the model captures both inter-bus dependencies and power grid topological characteristics, yielding the high-dimensional bus representation $\boldsymbol{X}_D$. Since the subsequent bus power flow computation no longer depends on other bus features or the system topology, each bus representation $\boldsymbol{X}_{D,i}$ can be regarded as the complete power flow state of bus $i$.

\subsection{Design of the Multi-Task Output}
\label{section 3.3}
To jointly predict bus and branch power flows, RMGL employs a multi-task output architecture. This architecture consists of two task-specific branches that perform feature downscaling and output prediction for buses and branches.

For bus power flow prediction, the high-dimensional bus representation $\boldsymbol{X}_D$ is mapped to outputs using a fully connected layer:
\begin{equation}
  {\widetilde {\boldsymbol{X}}_{out}} = {{\boldsymbol{X}}_D}{{\boldsymbol{W}}_{bus}}
  \label{eq11}
\end{equation}

Branch power flow prediction must satisfy two key physical principles. 1) According to (\ref{eq1}), branch power depends only on the states of its two terminal buses and itself, so the transformation must remain localized. 2) Both directions of branch flows must be predicted simultaneously to support physical consistency constraints.

For a branch $l$ connecting buses $i\rightarrow j$, the power prediction is formulated as:
\begin{equation}
  {{\boldsymbol{H}}_l} = \left( {{{\boldsymbol{X}}_{D,i}}\parallel {{\boldsymbol{X}}_{D,j}}\parallel \left( {{{\boldsymbol{X}}_{D,i}} - {{\boldsymbol{X}}_{D,j}}} \right)\parallel {{\boldsymbol{A}}_{ij}}{{\boldsymbol{W}}_A}} \right){{\boldsymbol{W}}_H}
  \label{eq12}
\end{equation}
\begin{equation}
  {\widetilde {\boldsymbol{H}}_{out,l}} = {{\boldsymbol{H}}_l}{{\boldsymbol{W}}_{branch}}
  \label{eq13}
\end{equation}
where ${{\boldsymbol{W}}_A} \in {\mathbb{R}^{3 \times {d_D}}}$ is the branch-level feature projection matrix. ${{\boldsymbol{W}}_H} \in {\mathbb{R}^{4{d_{D}} \times {d_{D}}}}$ is the branch feature extraction matrix, which derives representative features from both the terminal buses and the branch itself. ${{\boldsymbol{W}}_{branch}} \in {\mathbb{R}^{{d_D} \times 2}}$ is the projection matrix used to map the extracted features to the final power flow outputs. The above formulation is sensitive to the bus order, allowing directional prediction at both ends of the branch. This property lays the foundation for the physics-informed constraints introduced later.

\section{Training and Evaluation}
\label{section 4}
This section presents the training and evaluation strategy of the proposed model. \cref{section 4.1} introduces the LTS method for training data augmentation. \cref{section 4.2} analyzes the potential violations of physical laws in existing training strategies and develops a more comprehensive hybrid loss function. \cref{section 4.3} describes the evaluation metrics, focusing on the maximum error of each test sample.

\subsection{Training Sample Augmentation Based on LTS}
\label{section 4.1}

\begin{figure}[t]
\centerline{\includegraphics[width=1.0\columnwidth]{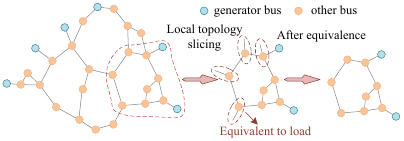}}
\vspace{-0.35cm}
\caption{Local topology slicing.}
\label{fig:LTS}
\vspace{-0.1cm}
\end{figure}

The topology of a power system can be represented as a graph $\cal G$. By extracting subgraphs ${\cal{G}}_{sub}$ of arbitrary scales from the complete topology, the distributions of bus and branch power flows within each subgraph still satisfy the power flow equations. Therefore, any local topology of the grid can be regarded as an independent sample for model learning.

As shown in \cref{fig:LTS}, the LTS method extracts subgraphs with different bus numbers and connection patterns, significantly enhancing both the quantity and diversity of training samples. The extraction follows two basic principles:
\begin{enumerate}[label=$\bullet$]
  \item Each subgraph must include at least one generator bus to ensure power balance.
  \item The power on boundary tie lines must be equivalently represented as load injections to maintain consistency with the original grid.
\end{enumerate}

The core procedure of LTS is as follows:
\begin{enumerate}
  \item Randomly select a starting bus and expand the subgraph based on the adjacency matrix until the target number of buses is reached.
  \item Replace the power of tie lines outside the subgraph with equivalent loads to preserve power flow consistency.
  \item Introduce power perturbations and branch outages to generate samples with diverse flow distributions.
\end{enumerate}

This sampling method, combining topological and power perturbations, can produce an almost unlimited number of training samples even from limited original data, thereby improving the structural coverage of the training set.

\subsection{Data-Physics Joint Driven Loss Function}
\label{section 4.2}
\begin{figure}[t]
\centerline{\includegraphics[width=0.8\columnwidth]{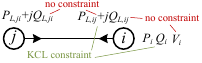}}
\caption{Model outputs and their physical constraints.}
\vspace{-0.35cm}
\label{fig:constraints}
\vspace{-0.1cm}
\end{figure}

Existing DL based PFA studies typically combine MSELoss with KCL constraints \cite{ref22}. However, within SaMPFA, relying solely on KCL is insufficient to ensure full physical consistency. As shown in \cref{fig:constraints}, KCL only enforces local power balance between buses and adjacent branches but does not constrain branch end differences or the relation between branch power and bus voltage. Therefore, additional physical mechanisms are incorporated to guide the model toward better adherence to physical laws.

To achieve both numerical accuracy and physical consistency, a hybrid loss function is defined as:
\begin{equation}
{\cal L} = {\varepsilon _{data}}{{\cal L}_{data}} + {\varepsilon _{phy}}{{\cal L}_{phy}}
\label{eq14}
\end{equation}
where ${\varepsilon _{data}}$ and ${\varepsilon _{phy}}$ are the weighting factors for the data-driven loss and physics-driven loss.

\subsubsection{Data-Driven Loss}
\label{section 4.2.1}
The data-driven term consists of weighted mean-squared errors at both bus and branch:
\begin{equation}
    {{\cal L}_{data}} = \frac{{{\varepsilon _N}}}{B}\sum\limits_b^B {{{\left\| {\widetilde {\boldsymbol{X}}_{out}^{(b)} - {\boldsymbol{X}}_{out}^{(b)}} \right\|}^2}} 
    + \frac{{{\varepsilon _E}}}{B}\sum\limits_b^B {{{\left\| {\widetilde {\boldsymbol{H}}_{out}^{(b)} - {\boldsymbol{H}}_{out}^{(b)}} \right\|}^2}} 
\label{eq15}
\end{equation}
where $B$ is the number of training samples, and $b$ is the sample index. ${\varepsilon _N}$ and ${\varepsilon _E}$ are the weights for bus and branch-level prediction errors.

\subsubsection{Physics-Driven Loss}
\label{section 4.2.2}
To improve the physical consistency of the model outputs, we introduce the KCL constraint, branch loss constraint, and angle reconstruction constraint into the loss function.

KCL constraint ${{\cal L}_{KCL}}$ minimizes bus unbalanced power to ensure compliance with Kirchhoff's Current Law:
\begin{equation}
  {{\cal L}_{KCL}} = \frac{1}{B}\sum\limits_b^B {\sum\limits_i^{{N^{(b)}}} {\left( {\Delta \widetilde P_i^{(b)} + \Delta \widetilde Q_i^{(b)}} \right)/{N^{(b)}}} }
  \label{eq16}
\end{equation}
\begin{equation}
  \left\{ \begin{array}{l}
\Delta {\widetilde P_i} = \left| {{{\widetilde P}_i} + \sum\limits_{j \in {\cal N}(i)} {\left( {{{\widetilde P}_{L,ij}} + \widetilde V_i^2{g_{m,ij}}} \right)} } \right|\\
\Delta {\widetilde Q_i} = \left| {{{\widetilde Q}_i} + \sum\limits_{j \in {\cal N}(i)} {\left( {{{\widetilde Q}_{L,ij}} - \widetilde V_i^2{b_{m,ij}}} \right)} } \right|
\end{array} \right.
\label{eq17}
\end{equation}
where ${N^{(b)}}$ is the number of buses in sample $b$. $\Delta \widetilde P_i^{(b)}$ and $\Delta \widetilde Q_i^{(b)}$ are the active and reactive power imbalances at bus $i$. ${\cal N}(i)$ is the set of neighbors of bus $i$. ${g_{m,ij}}$ and ${b_{m,ij}}$ are the shunt conductance and susceptance of branch $l$.

Branch loss constraint ${{\cal L}_{loss}}$ compares the predicted branch losses with the physical losses calculated from predicted voltages and currents to ensure power conservation:
\begin{equation}
  {\widetilde I_{ij}} = \sqrt {\widetilde P_{L,ij}^2 + \widetilde Q_{L,ij}^2} /{\widetilde V_i}
  \label{eq18}
\end{equation}
\begin{equation}
  {\widetilde P_{loss,ij}} = \left| {\widetilde I_{ij}^2{r_{L,ij}}} \right|,{\widetilde Q_{loss,ij}} = \left| {\widetilde I_{ij}^2{x_{L,ij}}} \right|
  \label{eq19}
\end{equation}
\begin{equation}
  {P_{loss,ij}} = \left| {{P_{L,ij}} + {P_{L,ji}}} \right|,{Q_{loss,ij}} = \left| {{Q_{L,ij}} + {Q_{L,ji}}} \right|
  \label{eq20}
\end{equation}
\begin{equation}
\begin{split}
    {{\cal L}_{loss}} = &\frac{1}{B}\sum\limits_b^B {\sum\limits_{l = (i,j)}^{2{E^{(b)}}} {\frac{{\left| {\widetilde P_{loss,ij}^{(b)} - P_{loss,ij}^{(b)}} \right|}}{{2{E^{(b)}}}}} } \\
     + &\frac{1}{B}\sum\limits_b^B {\sum\limits_{l = (i,j)}^{2{E^{(b)}}} {\frac{{\left| {\widetilde Q_{loss,ij}^{(b)} - Q_{loss,ij}^{(b)}} \right|}}{{2{E^{(b)}}}}} }
\end{split}
\label{eq21}
\end{equation}
where ${\widetilde I_{ij}}$ is the current computed from the predicted branch power flow. ${r_{L,ij}} + j{x_{L,ij}} = 1/({g_{L,ij}} + j{b_{L,ij}})$ is the branch impedance. ${P_{loss,ij}}$ and ${Q_{loss,ij}}$ are the predicted active and reactive power losses on branch $l$. ${E^{(b)}}$ is the number of branches in sample $b$.

Angle reconstruction constraint ${{\cal L}_{angle}}$ uses (\ref{eq4}) to estimate angle differences from predicted branch powers and voltages, guiding the model to learn their physical patterns:
\begin{equation}
  {{\cal L}_{angle}} = \frac{1}{B}\sum\limits_b^B {\sum\limits_{l = (i,j)}^{2{E^{(b)}}} {\left| {\widetilde \theta _{ij}^{(b)} - \theta _{ij}^{(b)}} \right|} /2{E^{(b)}}} 
  \label{eq22}
\end{equation}

Finally, the physics-driven loss is a weighted sum of the above three components, where ${\varepsilon _{KCL}}$, ${\varepsilon _{loss}}$, and ${\varepsilon _{angle}}$ are the weights for each constraint term.
\begin{equation}
  {{\cal L}_{phy}} = {\varepsilon _{KCL}}{{\cal L}_{KCL}} + {\varepsilon _{loss}}{{\cal L}_{loss}} + {\varepsilon _{angle}}{{\cal L}_{angle}}
  \label{eq23}
\end{equation}

\subsection{Evaluation Method Focused on Extreme Errors}
\label{section 4.3}
Most studies focus on average prediction errors across all buses and branches. However, in PFA, a single large deviation can compromise the reliability of the entire sample. Therefore, the maximum error per sample deserves greater attention.

For each test sample, the maximum errors of voltage magnitude, phase angle, branch power, and bus unbalanced power are denoted as $E_V$, $E_\theta$, $E_{SL}$, and $E_{\Delta S}$. For example, the average maximum error of voltage magnitude is defined as:
\begin{equation}
  {\overline E _V} = \frac{1}{{{B_t}}}\sum\limits_{b = 1}^{{B_t}} {E_V^{(b)}}  = \frac{1}{{{B_t}}}\sum\limits_{b = 1}^{{B_t}} {\mathop {\max }\limits_i \left| {\widetilde V_i^{(b)} - V_i^{(b)}} \right|} 
  \label{eq24}
\end{equation}
where $B_t$ is the number of samples in the test set.

In addition, we define an accuracy indicator $Acc$, representing the percentage of samples where all prediction errors fall below predefined thresholds:
\begin{equation}
  Acc = \frac{1}{{{B_t}}}\sum\limits_{b = 1}^{{B_t}} {\mathbb{I}\left( {E_V^{(b)} \le {\mu _V},E_{SL}^{(b)} \le {\mu _{SL}},E_{\Delta S}^{(b)} \le {\mu _{\Delta S}}} \right)} 
  \label{eq25}
\end{equation}
where $\mathbb{I}( \cdot )$ is the indicator function. The thresholds of $E_V$, $E_{SL}$, and $E_{\Delta S}$, denoted as $\mu_V$, $\mu_{SL}$, and $\mu_{\Delta S}$, are set to 0.01p.u., 10 MVA, and 10 MVA.

\section{Case Study}
\label{section 5}
This section demonstrates the effectiveness of the proposed method in both the IEEE 39-bus system and a real-world provincial power grid in China.

\subsection{Sample Generation}
\label{section 5.1}
\subsubsection{Case \Rmnum{1}: IEEE 39-Bus System}
\label{section 5.1.1}
The system consists of 39 buses and 46 branches. Initially, 8,760 baseline operating scenarios are generated by considering 0-3 generator outages or the addition of 1-2 new buses. From these, 5,000 scenarios are randomly selected, and a training set with 1 million samples is created using the LTS method described in \cref{section 4.1}. Furthermore, 3,000 additional baseline scenarios are selected, and 20\% fluctuations in bus power, along with 0-2 branch outages, are considered, generating a test set, a downscaled generalization set (Gen-D), and an upscaled generalization set (Gen-U), each with 100,000 samples. It is noted that, during the generation of Gen-D and Gen-U, the number of buses in the system is modified to 35 and 42, respectively. These configurations have not been seen in the training set, thereby serving to evaluate the model's generalization ability across different system scales.

\subsubsection{Case \Rmnum{2}: Real-World Provincial Power Grid}
\label{section 5.1.2}
Using 366 days of operational data from a provincial power grid, 8,784 baseline scenarios are constructed. The first 300 days of data are divided into training and testing sets with a 4:1 ratio, producing 500,000 and 100,000 samples, respectively, following the same procedure as Case \Rmnum{1}. The remaining 66 days form the generalization set (Gen set), generating about 150,000 samples. The bus counts of the training, testing, and generalization sets range from 300-687, 633-686, and 634-690, respectively. Within the Gen set, 25,944 samples contain buses unseen during training and are defined as Gen-U, while the remaining 124,065 samples constitute the normal generalization set (Gen-N).

\begin{figure}[t]
\centerline{\includegraphics[width=0.8\columnwidth]{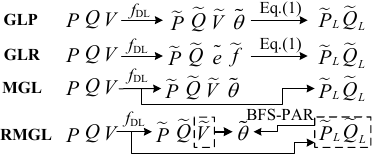}}
\vspace{-0.35cm}
\caption{Output designs of different models.}
\label{fig:output}
\end{figure}

\subsection{Training Strategy}
\label{seciton 5.2}
The model is built on the MindSpore framework, with training carried out on 32 Ascend 910B NPU cores, while inference is performed on 8 NPUs. Due to the large magnitude of angle reconstruction and branch loss in the early training stage, a two-stage training strategy is adopted. \textit{In the first stage}, only the data-driven loss and KCL constraint are used. The loss weights are set as: $\varepsilon_{data}=1$, $\varepsilon_{phy}=0.3$, $\varepsilon_{N}=1$, $\varepsilon_{E}=5$, $\varepsilon_{KCL}=0.5$. \textit{In the second stage}, angle reconstruction and branch loss terms are added, with weights $\varepsilon_{loss}=0.1$, $\varepsilon_{angle}=3$. The number of training epochs for the two stages is 200/1000 for Case \Rmnum{2}, and 200/200 for Case \Rmnum{2}.

\subsection{Verification of the Framework Design}
\label{section 5.3}
This section validates the advantages of the proposed framework through comparative experiments. Following the common output design strategies in existing DL-based PFA studies, three baseline models are constructed:
\begin{enumerate}[label=$\bullet$]
  \item GLP (Graph Learning under Polar coordinate): A single-task model that outputs the voltage magnitude and phase angle in polar coordinates. Its input and output design follows the same configuration as in \cite{ref11}.
  \item GLR: A single-task model that outputs the real and imaginary parts of bus voltages in rectangular coordinates, consistent with the output design in \cite{ref23}.
  \item MGL: A variant of RMGL that additionally includes the bus phase angle as both input and output variables.
\end{enumerate}

All three models are built under a fixed reference frame. Their output designs are illustrated in \cref{fig:output}, where $e$ and $f$ denote the real and imaginary parts of the bus voltages.

To ensure fairness, the four models are trained with identical training samples, input features, and feature extraction networks. Their performances in Case \Rmnum{1} and Case \Rmnum{2} are summarized in \cref{tab:performance1} and \cref{tab:performance2}.

\begin{table}[t]
    \footnotesize
    \vspace{-0.35cm}
    \setlength{\tabcolsep}{1mm}
     \centering
    \centering
    \caption{Performance Comparison in Case \Rmnum{1}}
    \label{tab:performance1}
    \vspace{-0.25cm}
    \begin{tabular}{>{\centering\arraybackslash}p{0.18\linewidth}cccccc} 
    \hline\hline
 \multirow{2}{*}{Dataset} & \multirow{2}{*}{Model} & Acc  & $E_V$                  & $E_\theta$           & $E_{SL}$  & $E_{\Delta S}$\\
                          &                        & (\%) &($\times 10^{-3}$p.u.)  & ($\times 10^{-2}$°)  & (MVA)     & (MVA) \\ \midrule
\multirow{4}{*}{Test set} & GLP   & 99.54 & \textbf{0.25} & 2.68 & 0.80  & 1.03 \\
                          & GLR   & 99.63 & \textbf{0.25} & 3.14 & 0.91  & 0.90 \\
                          & MGL   & 99.64 & 1.11 & 3.90 & 0.73  & 0.67 \\
                          & RMGL  & \textbf{99.87} & 0.62 & \textbf{1.88} & \textbf{0.43}  & \textbf{0.43} \\ \midrule
\multirow{4}{*}{Gen-D}    & GLP   & 98.74 & \textbf{0.62} & 5.87 & 1.28  & 1.39 \\
                          & GLR   & 98.79 & 0.66 & 6.68 & 1.50  & 1.31 \\
                          & MGL   & 98.53 & 1.40 & 6.26 & 1.21  & 1.09 \\
                          & RMGL  & \textbf{99.29} & 0.78 & \textbf{3.15} & \textbf{0.73}  & \textbf{0.69} \\ \midrule
\multirow{4}{*}{Gen-U}    & GLP   & 63.98 & 6.40 & 3.84 & 14.12 & 9.21 \\
                          & GLR   & 66.87 & 5.38 & 4.78 & 9.59  & 7.24 \\
                          & MGL   & 66.18 & 6.42 & 6.50 & 5.21  & 3.86 \\
                          & RMGL  & \textbf{82.42} & \textbf{3.38} & \textbf{3.48} & \textbf{3.93}  & \textbf{2.47} \\ \hline\hline
    \end{tabular}
\end{table}

\begin{table}[t]
    \footnotesize
    \vspace{-0.35cm}
    \setlength{\tabcolsep}{1mm}
     \centering
    \centering
    \caption{Performance Comparison in Case \Rmnum{2}}
    \label{tab:performance2}
    \vspace{-0.25cm}
    \begin{tabular}{>{\centering\arraybackslash}p{0.18\linewidth}cccccc} 
    \hline\hline
 \multirow{2}{*}{Dataset} & \multirow{2}{*}{Model} & Acc  & $E_V$                  & $E_\theta$           & $E_{SL}$  & $E_{\Delta S}$\\
                          &                        & (\%) &($\times 10^{-3}$p.u.)  & ($\times 10^{-2}$°)  & (MVA)     & (MVA) \\\midrule
\multirow{4}{*}{Test set} & GLP   & 50.72 & 2.14 & 66.42 & 22.94 & 3.65 \\
                          & GLR   & 47.09 & \textbf{2.13} & 72.14 & 33.49 & 9.30 \\
                          & MGL   & 96.07 & 4.75 & 14.43 & 2.94  & 4.01 \\
                          & RMGL  & \textbf{99.28} & 3.11 & \textbf{8.80}  & \textbf{1.65}  & \textbf{2.22} \\ \midrule
\multirow{4}{*}{\shortstack{Gen set\\(Gen-N \\\& Gen-U)}}       & GLP   & 46.66 & 2.81 & 75.89 & 25.93 & 3.40 \\
                          & GLR   & 39.59 & \textbf{2.25} & 88.89 & 39.02 & 9.38 \\
                          & MGL   & 90.73 & 5.88 & 19.25 & 3.73  & 4.58 \\
                          & RMGL  & \textbf{97.98} & 3.59 & \textbf{11.48} & \textbf{2.10}  & \textbf{2.59} \\ \hline\hline
    \end{tabular}
\end{table}

The comparison yields the following insights:
\begin{enumerate}
  \item When the bus angle is included in the output (MGL), the model struggles to learn absolute angles, especially in Case \Rmnum{2}. This difficulty also affects the learning of other variables.
  \item Compared with single-task models (GLP and GLP), the proposed multi-task framework effectively suppresses error amplification. As shown in \cref{tab:performance1}, GLP and GLP exhibit voltage and angle errors of about $6\times10^{-3}$p.u. and $4\times10^{-2}$° in Gen-U. These small errors are magnified by the power flow equations, leading to branch power errors up to 10 MVA. The problem becomes more severe in Case \Rmnum{2}, particularly in real-world networks with many low-impedance branches. By contrast, the proposed multi-task design reduces error propagation and significantly improves the accuracy of branch power prediction.
\end{enumerate}

\subsection{Effect of LTS on Scale Adaptability}
\label{section 5.4}
To evaluate the impact of the LTS on training effectiveness, a comparative experiment is conducted. In the sample generation, training samples are generated without using LTS, considering only bus power fluctuations and branch outages for data augmentation. This sample construction approach is similar to that in \cite{ref16,ref22}. The model is trained on 1 million samples and is denoted as RMGL-LTS. 

\begin{figure}[t]
\centerline{\includegraphics[width=1.0\columnwidth]{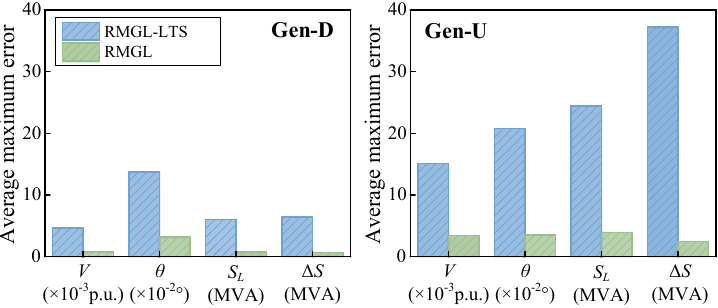}}
\vspace{-0.3cm}
\caption{Effect of LTS in Case \Rmnum{1}.}
\label{fig:case1LTS}
\end{figure}

As shown in \cref{fig:case1LTS}, the model trained without LTS exhibits significantly higher prediction errors in Case \Rmnum{1}, particularly in the Gen-U dataset. This is because grid expansion introduces notable shifts in graph characteristics, while the model without LTS is only exposed to systems containing 36-41 buses during training. The limited exposure to diverse topologies causes partial overfitting and poor generalization to unseen system scales.

\begin{figure}[t]
\centerline{\includegraphics[width=1.0\columnwidth]{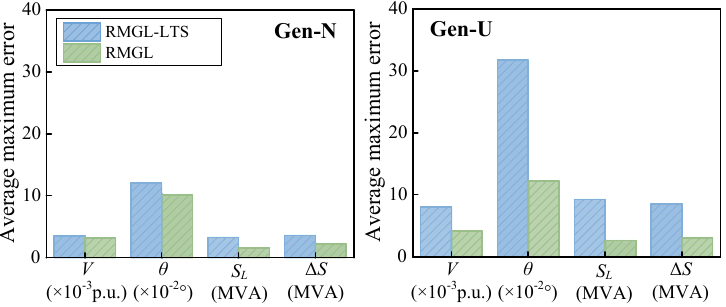}}
\vspace{-0.3cm}
\caption{Effect of LTS in Case \Rmnum{2}.}
\label{fig:case2LTS}
\end{figure}

In Case \Rmnum{2}, the same conclusion holds. Although the training set includes systems with 632-687 buses, allowing the model to learn some general power flow patterns across different scales, its topological diversity remains insufficient compared with the LTS-based training. As shown in \cref{fig:case2LTS}, the performance improvement from LTS in the Gen-N dataset is moderate; however, when the grid contains unseen structures, the difference becomes significant. LTS reduces the prediction errors of the four electrical quantities by 48.31\%, 61.50\%, 71.68\%, and 64.16\%, respectively, demonstrating that it substantially enhances model adaptability and generalization capability under system-scale variations.

\begin{figure}[t]
\centerline{\includegraphics[width=1.0\columnwidth]{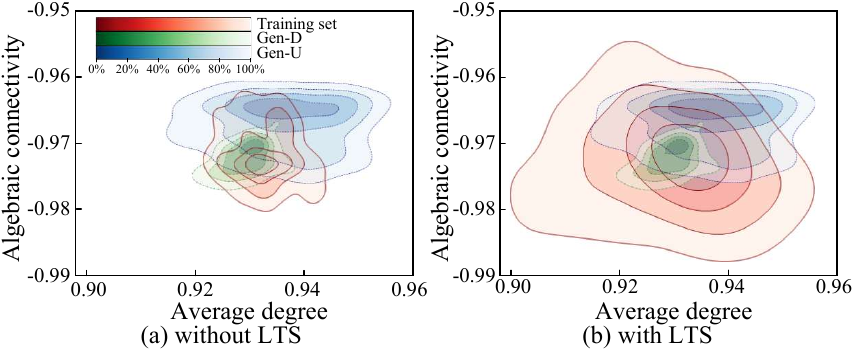}}
\vspace{-0.3cm}
\caption{Graph structure distribution of datasets.}
\label{fig:exp-LTS}
\end{figure}

To further investigate the influence of LTS on the distribution of graph features in training samples, the distributions of the training set, Gen-D, and Gen-U in Case \Rmnum{1} are visualized, as shown in \cref{fig:exp-LTS}. Without LTS, the training distribution is narrow, with large overlap between the training set and Gen-D but limited overlap with Gen-U, resulting in inferior performance on Gen-U. After introducing LTS, the training distribution expands significantly, fully covering Gen-D and most of Gen-U, thereby greatly improving model performance across both datasets.

\subsection{Analysis of Expansion Scenarios}
\label{section 5.5}

\begin{figure}[t]
\centerline{\includegraphics[width=0.9\columnwidth]{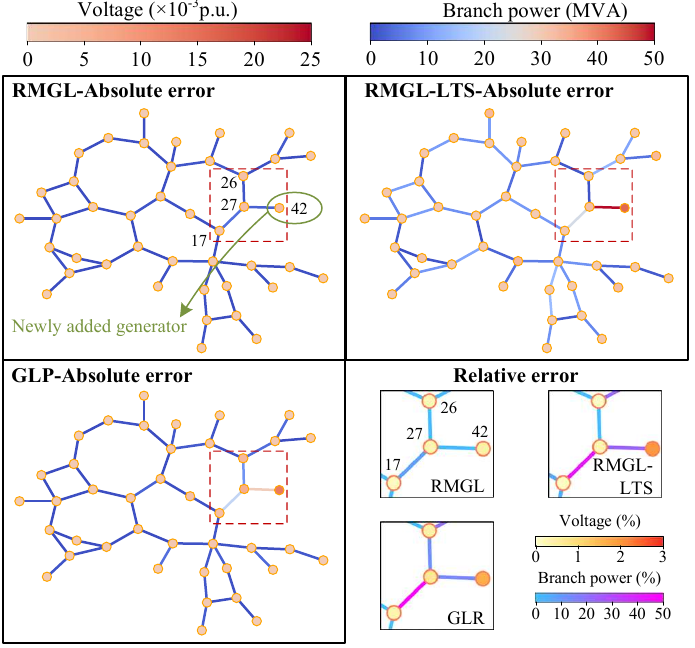}}
\vspace{-0.3cm}
\caption{Performance on an expansion sample in Case \Rmnum{1}.}
\label{fig:case1sample}
\end{figure}

To evaluate the adaptability of the proposed method to grid expansion, one expansion scenario from Case \Rmnum{1} is examined, where a new generator bus \#42 is connected to bus \#27. As shown in \cref{fig:case1sample}, when the system is expanded, RMGL-LTS and GLP perform poorly in the local area around the new bus, particularly in branch power prediction. Their maximum branch errors are 44.70 MVA and 27.63 MVA higher than the proposed model. For example, on branch \#17-\#27 with an actual power flow of 57.31 MVA, the prediction errors of RMGL, RMGL-LTS, and GLP are 8.08\%, 46.02\%, and 39.32\%, respectively. These results demonstrate that the proposed model exhibits stronger adaptability to topological changes.

From a theoretical perspective, RMGL-LTS is trained only on networks with 36-41 buses, which weakens its ability to generalize when the number of buses increases. Although GLP benefits from LTS and performs better than RMGL-LTS, its output design couples bus voltages and branch powers too tightly. As shown in \cref{fig:case1sample}, the relative errors of voltages at buses \#17 and \#27 are both below 0.70\%. However, due to error amplification in the branch flow equations, the power error on branch \#17-\#27 reaches 39.32\%. By decoupling bus voltages and branch powers, the proposed multi-task architecture achieves greater stability under system expansion.

\begin{figure}[t]
\centerline{\includegraphics[width=1.0\columnwidth]{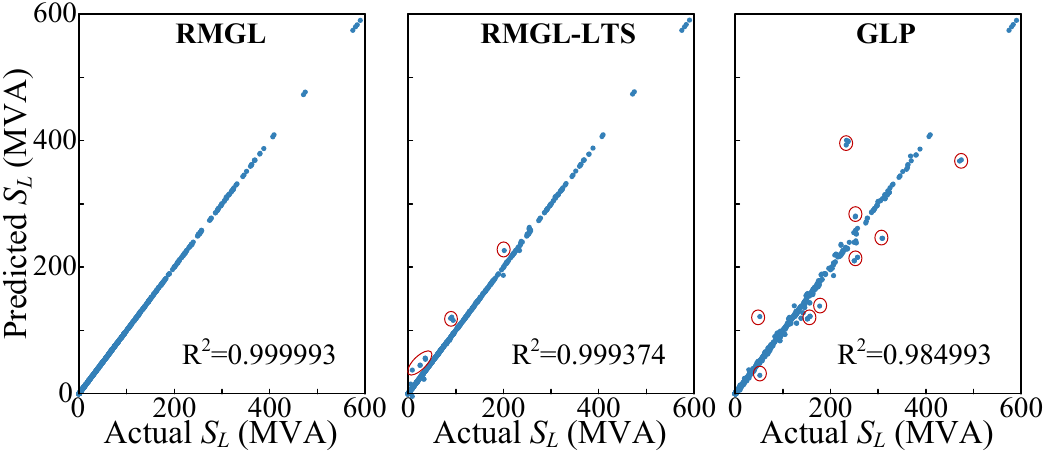}}
\vspace{-0.3cm}
\caption{Actual and predicted branch powers in an unseen scenario of Case \Rmnum{2}.}
\label{fig:case2error}
\end{figure}

The same test is conducted using a scenario from Case \Rmnum{2} that includes two newly added photovoltaic (PV) stations. The three models predict the power flows of all branches. As shown in \cref{fig:case2error}, the proposed model achieves the best performance, with predicted branch powers closely matching actual values across the system.

\begin{figure}[t]
\centerline{\includegraphics[width=0.9\columnwidth]{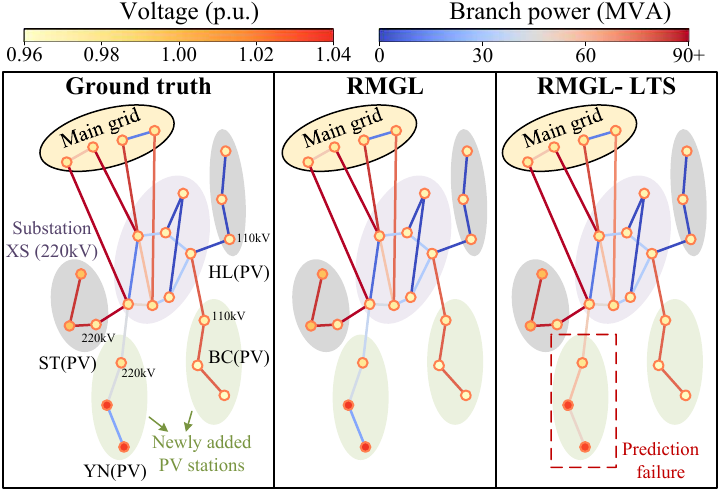}}
\vspace{-0.3cm}
\caption{Prediction results for the local area with newly added PV stations.}
\label{fig:case2expand}
\end{figure}

We also compare prediction accuracy in the expanded local topology. Since RMGL-LTS does not use LTS, it shows poor accuracy in unseen topologies. As shown in \cref{fig:case2expand} two new PV stations (\#YN and \#BC) are connected to different buses of the 220 kV substation \#XS. While RMGL-LTS correctly predicts the power of \#BC, it produces a large error at \#YN, predicting 49.82 MVA instead of the actual 20.35 MVA. In contrast, the proposed model provides accurate predictions in this region, highlighting the role of LTS in improving adaptability to real-world grid expansion.

\begin{figure}[t]
\centerline{\includegraphics[width=0.9\columnwidth]{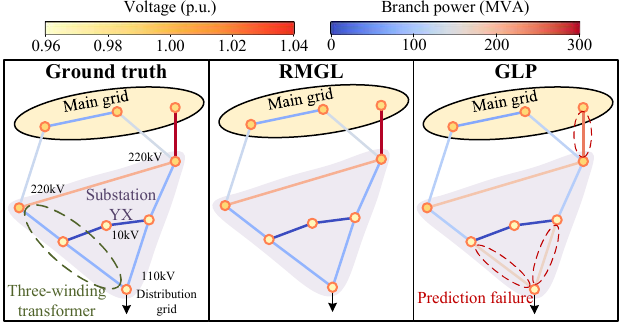}}
\vspace{-0.3cm}
\caption{Prediction results for lines inside the substation.}
\label{fig:case2substation}
\end{figure}

Compared with GLP, the proposed model shows clear advantages in predicting lines within the substation. As shown in \cref{fig:case2substation}, both models predict voltages accurately in the substation. However, due to the low impedance of the medium-voltage side of the three-winding transformer, STP suffers severe error amplification, leading to a branch error as large as 90 MVA. This error propagates through the KCL constraint, reducing accuracy in adjacent branches. By decoupling bus voltage and branch power outputs, the proposed design prevents error propagation and significantly improves branch power prediction reliability.

\subsection{Hybrid Loss Function Enhances Physical Consistency}
\label{section 5.6}
This section evaluates the effect of branch loss and angle difference constraints on the physical consistency of model outputs through ablation experiments. Four test settings are conducted in both case studies, and the results are reported in \cref{tab:phy1} and \cref{tab:phy2}. With the branch loss constraint ${{\cal L}_{loss}}$, the physical consistency of active and reactive power between branch terminals improves significantly, reducing branch loss errors by 61.91\% and 45.14\% in the two systems. The angle difference constraint ${{\cal L}_{angle}}$ effectively reduces angle errors, which improves the accuracy of branch power prediction and decreases errors in the BFS-PAR recovery process. When both constraints are applied together, the model combines the benefits of each, leading to tighter alignment between outputs and physical equations.

\begin{table}[t]
    \footnotesize
    \vspace{-0.35cm}
    \setlength{\tabcolsep}{1mm}
     \centering
    \centering
    \caption{Physical Constraint Violations in Case \Rmnum{1}}
    \label{tab:phy1}
    \vspace{-0.25cm}
    \begin{tabular}{cccccccc} 
    \hline\hline
 \multirow{3}{*}{${{\cal L}_{loss}}$} & \multirow{3}{*}{${{\cal L}_{angle}}$} & \multicolumn{3}{c}{Test set (36-41 buses)}     & \multicolumn{3}{c}{Gen set (35\&42 buses)} \\ \cline{3-8} 
                          &  & $P_{loss}$ &$Q_{loss}$  & $\Delta \theta$  & $P_{loss}$ &$Q_{loss}$  & $\Delta \theta$ \\
                          &  & (MW) &(MVar)  & ($\times 10^{2}$°)  & (MW) &(MVar)  & ($\times 10^{2}$°) \\ \midrule
$\times$ & $\times$ & 0.07 & 0.26 & 1.28 & 0.42 & 1.22 & 3.31 \\
$\surd$ & $\times$ & 0.06 & 0.22 & 1.08 & 0.09 & 0.58 & 3.16 \\
$\times$ & $\surd$ & 0.06 & 0.23 & 0.91 & 0.34 & 0.90 & 2.02 \\
$\surd$ & $\surd$ & 0.06 & 0.21 & 0.94 & 0.09 & 0.61 & 1.98 \\ \hline \hline
    \end{tabular}
\end{table}

\begin{table}[t]
    \footnotesize
    \vspace{-0.35cm}
    \setlength{\tabcolsep}{1mm}
     \centering
    \centering
    \caption{Physical Constraint Violations in Case \Rmnum{2}}
    \label{tab:phy2}
    \vspace{-0.25cm}
    \begin{tabular}{cccccccc} 
    \hline\hline
 \multirow{3}{*}{${{\cal L}_{loss}}$} & \multirow{3}{*}{${{\cal L}_{angle}}$} & \multicolumn{3}{c}{Test set}     & \multicolumn{3}{c}{Gen set} \\\cline{3-8} 
                          &  & $P_{loss}$ &$Q_{loss}$  & $\Delta \theta$  & $P_{loss}$ &$Q_{loss}$  & $\Delta \theta$ \\
                          &  & (MW) &(MVar)  & ($\times 10^{2}$°)  & (MW) &(MVar)  & ($\times 10^{2}$°) \\ \midrule
$\times$ & $\times$ & 0.92 & 0.76 & 8.84 & 1.18 & 0.91 & 11.97 \\
$\surd$ & $\times$ & 0.40 & 0.51 & 7.43 & 0.44 & 0.74 & 10.64 \\
$\times$ & $\surd$ & 0.79 & 0.59 & 4.13 & 0.83 & 0.75 & 7.37  \\
$\surd$ & $\surd$ & 0.44 & 0.52 & 4.28 & 0.46 & 0.68 & 6.34  \\\hline \hline
    \end{tabular}
\end{table}

\subsection{Model Initialization Improves NR Convergence}
\label{section 5.7}
Finally, the effect of model outputs on the convergence of the Newton-Raphson (NR) method is investigated. By increasing system load levels, 20,000 test samples are generated in both systems. The NR method is first initialized with a flat start, and the convergence status, average iteration counts of all samples, and runtime are recorded (with a maximum of 200 iterations). The proposed model is then used to infer bus voltages and angles, which are employed as the initial state of NR, and the same data are recorded.

As shown in \cref{tab:conv}, initializing NR with DL model outputs significantly improves convergence. In Case II, the convergence rate increases by 40.16\%. The number of NR iterations is also greatly reduced, leading to runtime improvements of 16.18\% and 20.13\% in the two systems. Considering DL inference alone, the proposed model achieves speedups of 535$\times$ and 233$\times$ compared with NR, making it a superior choice for rapid online analysis of large sample sets.

\begin{table}[t]
    \footnotesize
    \vspace{-0.35cm}
    \setlength{\tabcolsep}{1mm}
     \centering
    \centering
    \caption{Impact of Model initialization on NR Convergence}
    \label{tab:conv}
    \vspace{-0.25cm}
    \begin{tabular}{cccccc} 
    \hline\hline
Case                & Method & Convergence rate(\%) & Iterations & DL time(s) & NR time(s) \\ \midrule
\multirow{2}{*}{\Rmnum{1}}  & NR     & 100\%                   & 5.313      & /             & 578.21        \\
                    & DL+NR  & 100\%                   & 2.001      & 1.08          & 484.65        \\
\multirow{2}{*}{\Rmnum{2}} & NR     & 54.19\%                 & 78.821     & /             & 4813.43       \\
                    & DL+NR  & 94.35\%                 & 9.642      & 20.70         & 3844.44       \\\hline \hline
    \end{tabular}
\end{table}

\section{Conclusion}
\label{section 6}
This paper presents a novel SaMPFA framework to enhance the applicability of DL models in real-world power systems. By optimizing both model architecture and training strategy, the proposed approach achieves strong adaptability and accuracy across varying system scales and complex topologies. Experiments on the IEEE 39-bus system and a real provincial grid lead to the following conclusions:

1) SaMPFA achieves superior adaptability and generalization under system-scale variations. Accuracy improves by 0.27\%/8.67\% for seen and unseen scales in the IEEE 39-bus system, and by 34.65\%/38.99\% in the real-world system, confirming its scalability.

2) LTS effectively enhances the diversity of graph features, improving model robustness in grid expansion scenarios. The RMGL model exhibits higher stability and accuracy in branch power prediction, especially in real-world systems.

3) Using RMGL predictions as initialization significantly accelerates numerical power flow convergence, reducing iterations by 37.62\% in the IEEE 39-bus system and improving convergence rate by 40.16\% in the real-world system.

This study represents a first step toward building a general PFA model adaptable to diverse power systems. Future work will extend the SaMPFA framework for multi-system joint training and rapid transfer to unseen systems.

\end{document}